\documentclass{article}

\PassOptionsToPackage{numbers}{natbib}
\usepackage[preprint]{nips_2018}

\usepackage[utf8]{inputenc} % allow utf-8 input
\usepackage[T1]{fontenc}    % use 8-bit T1 fonts
\usepackage{hyperref}       % hyperlinks
\usepackage{url}            % simple URL typesetting
\usepackage{booktabs}       % professional-quality tables
\usepackage{amsfonts}       % blackboard math symbols
\usepackage{nicefrac}       % compact symbols for 1/2, etc.
\usepackage{microtype}      % microtypography
\usepackage{graphicx}
\usepackage{caption}
\usepackage{subcaption}
\usepackage{placeins}
\usepackage{amsmath}
\usepackage{algorithmicx}
\usepackage{algorithm}
\usepackage{algpseudocode}

\DeclareMathOperator*{\argmax}{arg\,max}

\title{Combining Model-Free Q-Ensembles and Model-Based Approaches for Informed Exploration}

\author{
  Sreecharan Sankaranarayanan$^{*}$\\
  Language Technologies Institute\\
  Carnegie Mellon University\\
  Pittsburgh, PA 15213\\
  \texttt{sreechas@cs.cmu.edu} \\
  \And
  Raghuram Mandyam Annasamy$^{*}$\\
  Language Technologies Institute\\
  Carnegie Mellon University\\
  Pittsburgh, PA 15213\\
  \texttt{rannasam@andrew.cmu.edu} \\
  \And
  Katia Sycara\\
  Robotics Institute\\
  Carnegie Mellon University\\
  Pittsburgh, PA 15213\\
  \texttt{katia@cs.cmu.edu} \\
  \And
  Carolyn Penstein Rosé\\
  Language Technologies Institute\\
  Carnegie Mellon University\\
  Pittsburgh, PA 15213\\
  \texttt{cprose@cs.cmu.edu} \\
}

\begin{document}

\maketitle

\begin{abstract}
Q-Ensembles are a model-free approach where input images are fed into different Q-networks and exploration is driven by the assumption that uncertainty is proportional to the variance of the output Q-values obtained. They have been shown to perform relatively well compared to other exploration strategies. Further, model-based approaches, such as encoder-decoder models have been used successfully for next frame prediction given previous frames. This paper proposes to integrate the model-free Q-ensembles and model-based approaches with the hope of compounding the benefits of both and achieving superior exploration as a result. Results show that a model-based trajectory memory approach when combined with Q-ensembles produces superior performance when compared to only using Q-ensembles.
\end{abstract}

\section{Introduction and Related Work}
Quantifying predictive uncertainty is a problem that has started to receive a lot of attention as Deep Neural Networks achieve state-of-the-art performance \cite{lecun2015deep} in a wide variety of domains such as computer vision \cite{krizhevsky2012imagenet}, speech recognition \cite{hinton2012deep}, natural language processing \cite{mikolov2013efficient} and bio-informatics \cite{zhou2015predicting, alipanahi2015predicting} but have continued to produce overconfident estimates. These overconfident estimates can be detrimental or even harmful for practical applications \cite{amodei2016concrete}. Therefore, quantifying predictive uncertainties aside from just the accuracy of networks is an important problem. The contribution of our work is to combine an encoder-decoder model-based architecture and trajectory memory with the model-free Q-ensemble approach for the purpose of uncertainty estimation which in the context of reinforcement learning is exploration.

\subsection{Neural Network Ensembles for Uncertainty Prediction}
Current approaches to quantifying uncertainty have mostly been Bayesian where a prior distribution is specified over the parameters of the Neural Network and then using the training data, the computed posterior distribution over the parameters is used to calculate the uncertainty \cite{bernardo2009bayesian}. Since this form of Bayesian inference is computationally intractable, approaches have ranged from Laplace Approximation \cite{mackay1992bayesian}, Markov Chain Monte Carlo methods \cite{neal2012bayesian} to Variational Bayesian Inference methods \cite{blundell2015weight, graves2011practical, louizos2016structured}. These methods however suffer from issues due to bounds of computational power and over-reliance on the correctness of the prior probability distribution over the parameters. Having priors of convenience can in fact lead to unreasonable uncertainty estimates \cite{rasmussen2005healing}. In order to overcome these challenges and produce a more robust uncertainty estimate, Lakshminarayanan et al. \cite{lakshminarayanan2017simple} proposed using an ensemble of Neural Networks trained under a defined scoring rule. This approach when compared to Bayesian approaches is much simpler, has parallelization advantages and achieves state-of-the-art or better performance. State-of-the-art performance before this was achieved by MC-dropout which can also in essence be considered an ensemble approach where the predictions are averaged over an ensemble of Neural Networks with parameter sharing \cite{srivastava2014dropout}.
\subsection{Model-Free Q-Ensembles}
This ensemble approach for uncertainty estimation in Neural Networks motivated its use for uncertainty estimation for exploration in the case of Deep Reinforcement Learning in the form of Q-Ensembles \cite{chen2018ucb}. Specifically, an ensemble voting algorithm is proposed where the agent takes action based on a majority vote over the Q-ensemble. The exploration strategy described uses the estimate of the confidence interval to then optimistically explore in the direction of the largest confidence interval (highest uncertainty). This approach was demonstrated to improve significantly over an Atari benchmark. The Q-Ensemble approach is an example of a model-free reinforcement learning approach where we do not need to infer the environment in the learning process. Model-free approaches are however generally high in sample complexity. Sampling from a learned model of the environment can help us mitigate this problem.
\subsection{Model-Based Encoder-Decoder Approach to Next-Frame Prediction}
Model-based approaches in the Atari environment (Arcade Learning Environment \cite{bellemare2013arcade}) have been successfully used in the past for the problem of next-frame prediction. The game environments are high in complexity with the frames themselves being high-dimensional, involving tens of objects that are being controlled directly or indirectly by agent actions and cases of objects leaving and entering the frame. Oh et al. \cite{oh2015action} described an encoder-decoder model that produces visually-realistic next frames which can then be used as control for action-conditionals in the game. They further described an informed exploration approach called \textit{trajectory memory} that follows the $\epsilon$-greedy strategy but leads to the frame that was least visited in the last n-steps instead of random exploration.
\subsection{Need for Combining Model-Free and Model-Based Approaches}
In order to address the issues related to modeling of the latent space, we propose taking advantage of a combination of the model-free and model-based approaches described so far. Typically, model-free approaches are very effective at learning complex policies but convergence might required millions of trials and could lead to globally sub-optimal local minima \cite{deisenroth2013survey}. On the other hand, model-based approaches have the theoretical benefit of being able to generalize to new tasks better and reduce the number of trials required significantly \cite {deisenroth2015gaussian, levine2014learning} but require an environment which is either engineered or learned well to achieve this generalization. Another advantage of the model-free approach is that it can model arbitrarily complex unknown dynamics better but is substantially less sample efficient as indicated earlier. Prior attempts at combining the two approaches while retaining the relative advantages have been met with some success \cite{bansal2017mbmf, chebotar2017combining, leibfried2016deep}. We will therefore be performing a combination of an encoder-decoder model and one that uses the informed trajectory memory exploration strategy that was proposed by Oh et al. \cite{oh2015action} with the Q-ensemble and report the results.\\
\\
Section 2 describes the two methods of combining model-based methods and model-free q-ensembles for exploration that we implemented. Section 3 details our experimental setup. Section 4 details the results we obtained and provides a discussion about the methods we attempted based on our experiments and related work. Section 5 concludes the paper with pointers about work we intend to do in the future.
\section{Methods}
\begin{figure}[ht!]
\centering
\includegraphics[width =\linewidth]{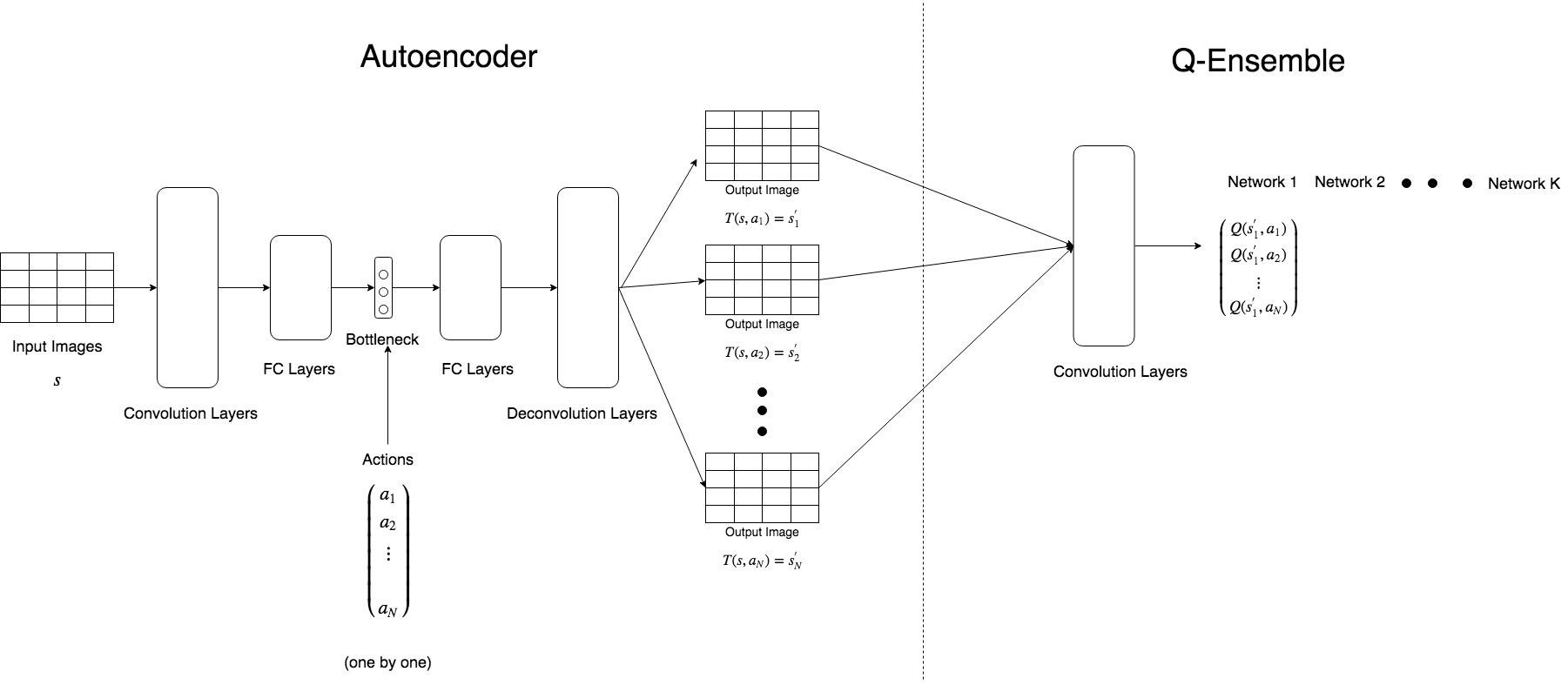}
\caption{Combination of Auto-encoder and Q-Ensemble}
\label{fig4}
\end{figure}
\begin{figure}[ht!]
\centering
\begin{subfigure}{.25\textwidth}
  \centering
  \includegraphics[width=0.95\linewidth]{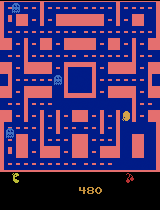}
  \caption{Ground Truth}
\end{subfigure}%
\begin{subfigure}{.25\textwidth}
  \centering
  \includegraphics[width=0.95\linewidth]{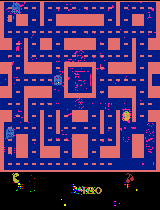}
  \caption{Predicted}
\end{subfigure}%
\begin{subfigure}{.25\textwidth}
  \centering
  \includegraphics[width=0.95\linewidth]{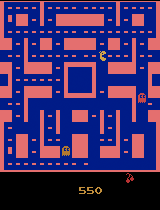}
  \caption{Ground Truth}
\end{subfigure}%
\begin{subfigure}{.25\textwidth}
  \centering
  \includegraphics[width=0.95\linewidth]{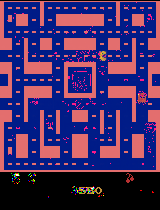}
  \caption{Predicted}
\end{subfigure}
\caption{Side-by-side Comparison of Ground Truth and Predicted Next Frames by Model-based Encoder-Decoder Approach at 744000 and 746000 Iterations Respectively.}
\label{fig1}
\end{figure}

\begin{figure}[ht!]
\centering
\includegraphics[height = 7cm]{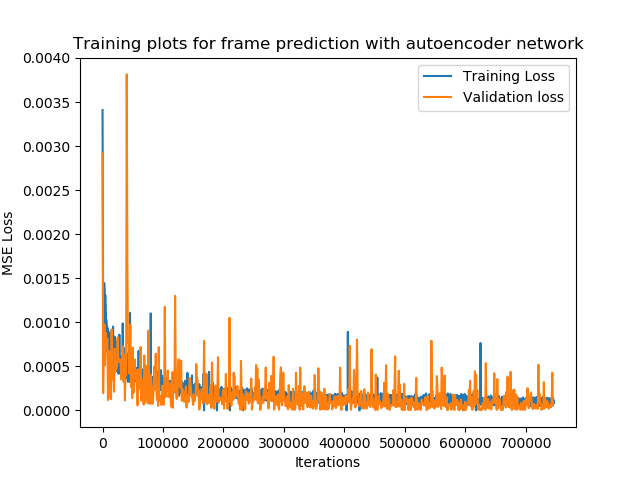}
\caption{Training and validation loss for frame prediction using an auto-encoder}
\label{fig3a}
\end{figure}

\begin{figure}[ht!]
\centering
\includegraphics[height = 7cm]{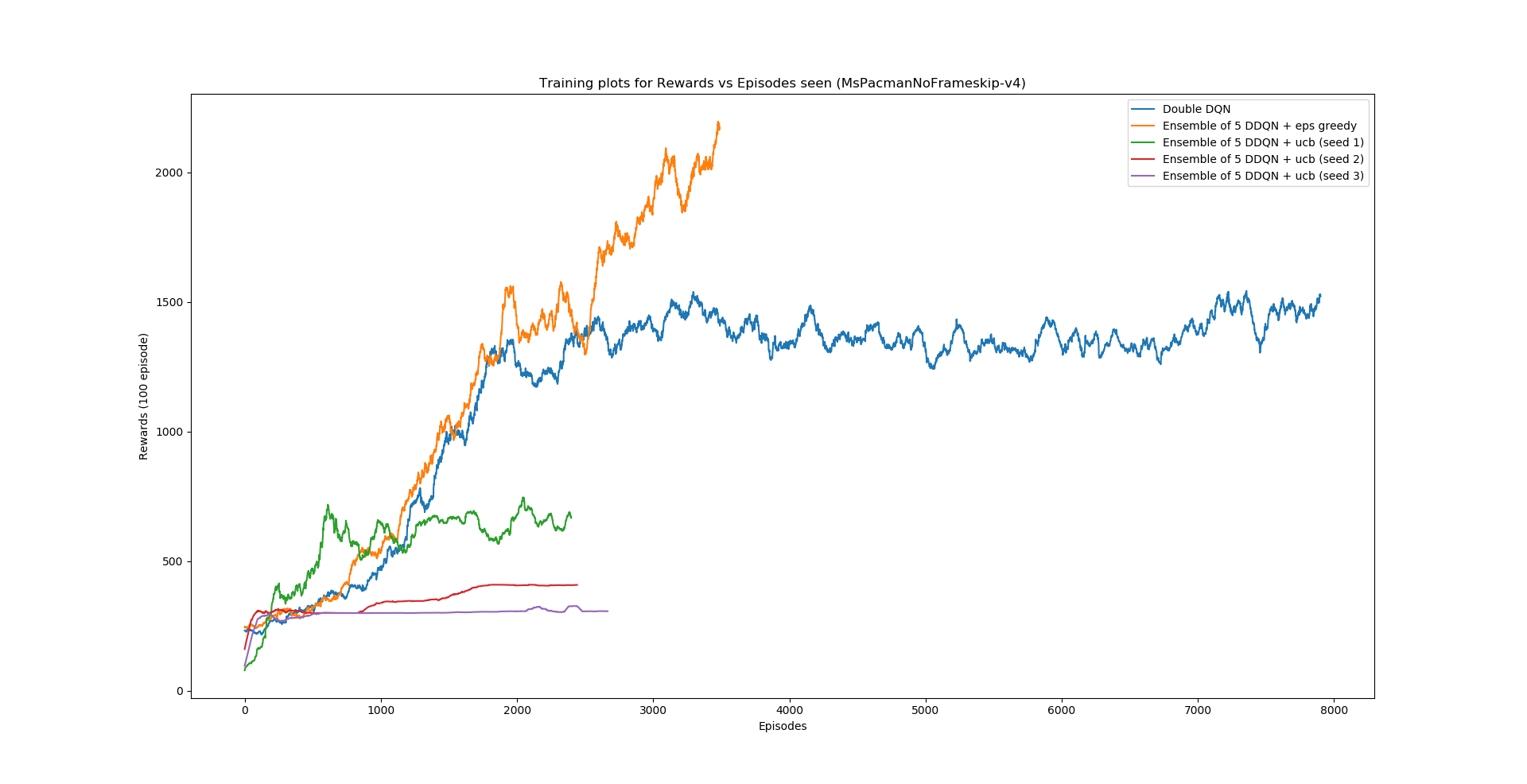}
\caption{DDQN and DDQN Ensemble using $\epsilon$-greedy and UCB approaches}
\label{fig3b}
\end{figure}
Figure \ref{fig4} visually represents the combination of the model-based and model-free Q-ensemble approaches to guide exploration. The model-based encoder-decoder is first used to predict next frames over all possible actions. Given each of those actions, now the Q-values associated with them are predicted using the Q-ensemble and the variance is used to drive exploration. This is somewhat similar to the model predictive control (MPC) framework \cite{garcia1989model} (but instead of planning over the action-tree, we simply repeat each action multiple times). Further details about the methods are provided in the following sections.
\subsection{Method 1 - Feeding Auto-Encoder Images Directly To Q-Ensemble}
In this method, we use the auto-encoder model to generate next-step frames for all of the actions and feed in these predicted frames to a Q-ensemble. We use the variance in the predicted Q-values (for each predicted frame) to estimate how likely this state has been visited during Q-learning (visit frequency). So to drive exploration, we need to simply pick those actions for which there is high variance. Given a well trained auto-encoder model, we can unroll many steps of predicted frames and select paths with high-variance.

Figure \ref{fig4} shows the architecture first of the encoder-decoder model which is composed of encoding layers that extract spatio-temporal features from the input frames, action-conditional transformation layers that then transform the encoded features into a next-frame prediction by using action variables as an additional input, finally followed by decoding layers that map the high level features back onto the pixel space.
\subsubsection{Encoding, Action-Conditional Transformation, Decoding}
Similar to \cite{oh2015action,leibfried2016deep}, the convolutional layers use a feedforward encoding that takes a concatenated set of previous images and extracts spatio-temporal features from them. The convolutional layers are essentially a functional transformation from the pixel space to a high level feature vector space by passing through multiple convolutional layers followed by a fully-connected layer at the end, each of which is followed by a non-linearity. The encoded feature vector is therefore - 
\[
h^{enc}_{t} = Conv(x_{t-m+1:t})
\]
where
$x_{t-m+1:t} \in R^{(m x c) x h x w} $ denotes $m$ frames of $h x w$ pixels with $c$ color channels at time $t$. 

The encoded feature vector is now transformed using multiplicative interactions with the control variables. Ideally, the transformation would looks as follows - \[
h^{dec}_{t,i} = \sum_{j,l}{W_{ijl}h_{t,j}^{enc}a_{t,l} + b_i}
\]
where $h^{dec}_t$ is the action transformed feature, $a_t$ is the action-vector at time t, W is a 3-way tensor weight and b is the bias term. However, computing the 3-way tensor is not scalable but allows the architecture to model different transformations for different actions as has been demonstrated in prior work \cite{taylor2009factored, sutskever2011generating, memisevic2013learning}. Therefore, an approximation of the 3-way tensor is used instead - \\
\[
h_t^{dec} = W^{dec}((W^{enc}h_t^{enc}) * (W^a a_t)) + b
\]

De-convolutions using CNNs perform the inverse operation of convolutions, transforming the 1 x 1 spatial regions onto d x d using de-convolutional kernels. In the architecture we implemented, the De-convolutions was performed as follows - \\
\[
x_{t+1} = Deconv(Reshape(h^{dec}))
\]
where Reshape is a fully connected layer and Deconv consists of multiple deconvolution layers.
\subsubsection{Modifying Q-Ensemble}
For each state-action pair, the encoder-decoder model produces an image. We now need to determine, which of these we are most uncertain about, in order to explore in that direction. For this purpose, the images are passed to several Q-networks. Each Q-network provides a distribution over actions and therefore outputs corresponding to the number of actions (9 in Pacman). The total number of outputs therefore will be number of networks (which in our case was 5) times the number of actions. A variance metric is then calculated over all of these outputs to explore in the direction of highest variance.
\[
\text{Uncertainty} (s_t) = 
\frac{1}{N} \sum_{j=1}^{N}\sum_{i=1}^{K}\big[{Q_i(s_t, a_j) - \frac{1}{K} \sum_{i=1}^{K}{Q_i(s_t, a_j)}}\big]^2
\]
\[
\text{Action} (s_t) = \argmax_{a}\{\text{Uncertainty} (s_{t+1}) \text{ where } s_{t+1} = T(s_t, a) \; \forall a \}
\]

where $N$ is the number of actions, $K$ is the number of Q-networks (ensemble), $T(s_t, a)$ is the transition function (frame prediction model) and Action$(s_t)$ denotes the action with highest uncertainty. Another way of estimating the variance can be, (converts Q-estimates to Value estimates of $s_t$) 
\[
\text{Uncertainty} (s_t) = 
\sum_{i=1}^{K} \big[ \max_{j=1}^{N}{Q_i(s_t, a_j)}  - \dfrac{1}{K} \sum_{i=1}^{K}\max_{j=1}^{N}{Q_i(s_t, a_j)} \big] ^2
\]

These methods of exploration can be combined with other common exploration strategies like $\epsilon$-greedy where instead of selecting random actions, we can select the action with highest uncertainty. 

\subsection{Method 2 - Trajectory Memory and Q-Ensembles}
The trajectory memory method as described in \cite{oh2015action} uses an $\epsilon$-greedy exploration policy. The \textit{trajectory memory} is used to measure the similarity between the predicted frame and the most recent $d$ frames is calculated to give the estimated visit frequency. We use the same settings as described in \cite{oh2015action}; $d=20$, $\delta=50$ and $\sigma=100$.
\[
	n_D(s_t) = \sum_{i=1}^{d} k(s_t, s_i); \text{ where } k(x, y) = \exp\big(-\frac{1}{\sigma}\sum_{j} \min(\max((x_j - y_j)^2 - \delta, 0), 1)\big)
\]

We can combine this exploration method with the variance in Q-ensembles to yield a more informed exploration strategy. This is similar to the previous method except that the uncertainty is calculated on the current frame $(s_t)$ instead of the predicted frames ($s'_{t+1}$). The predicted frames are used only to estimate their visit frequency. The algorithm used to select actions using the combined strategies is described in Algorithm ~\ref{Exploration}.

\begin{algorithm}
\caption{Combined Exploration Strategy\label{Exploration}}
\begin{algorithmic}[1]
\Procedure{action\_selector}{current state: $s_t$}
\State \textit{Init} $\epsilon = 1.0$
\State \textbf{for each action $a$}
\State $\qquad \text{Generate predicted next state using auto-encoder: } s'_{t+1} = T(s_t, a) $
\State $\qquad \text{Estimate visit frequency: } n_D(s'_{t+1})$
\State $\qquad \text{UCB like estimate: } \mu_a + \lambda \sigma_a$

$\qquad \text{where } \mu_a = \frac{1}{K} \sum_i^k Q_i(s_t, a)$ and $\sigma_a = \sqrt[]{\frac{1}{K} \sum_i^k ( Q_i(s_t, a) - \mu_a\big) ^ 2}$

\State $\qquad \text{Set } score(a) = \mu_a + \lambda \sigma_a - \epsilon \; n_D(s'_{t+1})$

\State action = $\argmax_a score(a)$
\State Decay, $\epsilon = \epsilon/\text{decay factor}$

\EndProcedure
\end{algorithmic}
\end{algorithm}
\section{Experimental Setup}
All our experiments are run on the Ms. Pacman Atari environment \cite{bellemare2013arcade} where exploration is challenging and it is important to achieve better scores. Below are the experimental settings and hyper-parameters used.
\begin{itemize}
\item \textit{Network Architecture}: For the Q-ensembles, we train 5 different Q-networks and each one uses a standard DQN architecture: Conv(32, 8, 4), Conv(64, 4, 2), Conv(64, 3, 1) and Linear(256) with ReLU activations throughout. The auto-encoder frame prediction model is the same as the one used in \cite{oh2015action}. 
\item \textit{Optimization}: For training the Q-Ensemble, we use Adam with a learning rate = 0.0001, weight decay = 0 and gradient norm clipped at 10 for every layer.
\item \textit{Training details:} Batch size = 32, Training Frequency = 4 (train every 4 frames), Discount Factor (gamma) = 0.99, Size of Replay Memory = 10000, Target network sync frequency = 1000 (fully replaced with the weights from training network)
\item \textit{Frame Preprocessing:} We use the same frame processing technique as used by \cite{mnih2015human} (frame skipping, max over 4 frames, RGB to gray-scale).
\item \textit{Exploration:} For $\epsilon$-greedy based strategies, Initial Epsilon = 1.0, Final Epsilon = 0.01, Exploration Timesteps = 1000000. UCB like strategy uses $\lambda$ = \{1.0 0.1, 0.01, 0.001\}

\end{itemize}

\section{Results and Discussion}
\subsection{Separately Training Auto-Encoder and Q-Ensemble}
We discuss about the results of training the auto-encoder and Q-ensemble models separately as reported in \cite{oh2015action} and \cite{chen2018ucb} respectively.
\paragraph{Model-Based Approach}
Results from the next-frame prediction done by Oh et al.'s model-based approach \cite{oh2015action} are shown in the figures. Figure \ref{fig1} shows ground truth next frames and predicted next frames produced by the encoder-decoder model side-by-side. Figure \ref{fig3a} further shows the training and validation losses over 800000 iterations.

\paragraph{Model-Free Q-Ensemble}
Figure \ref{fig3b} shows the results of training different Q-Ensemble methods and the standard Double DQN implementation. Due to the inherent slowness in training ensembles all of the models were not trained for the full 8000 epochs but they have been trained sufficiently to say that the ensemble with $\epsilon$-greedy outperforms the UCB and standard DDQN approaches which replicates the results reported by Chen et al. \cite{chen2018ucb}. These Q-ensemble models with different exploration strategies will serve as our baseline to beat.

\subsection{Method 1 - Combining Encoder-Decoder Model and Q-Ensemble}
\begin{figure}[ht]
\centering
\begin{subfigure}{0.6\textwidth}
  \centering
  \includegraphics[width=\textwidth]{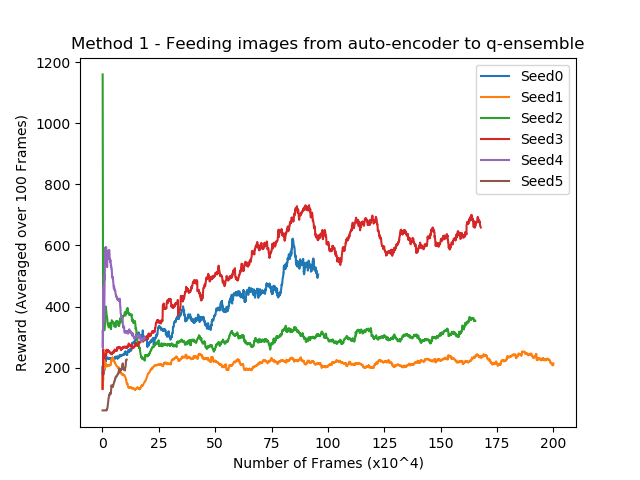}
  \label{fig:sub1}
  \caption{Average Rewards (Every 100 episodes) (Training)}
\end{subfigure}%

\begin{subfigure}{0.6\textwidth}
  \centering
  \includegraphics[width=\textwidth]{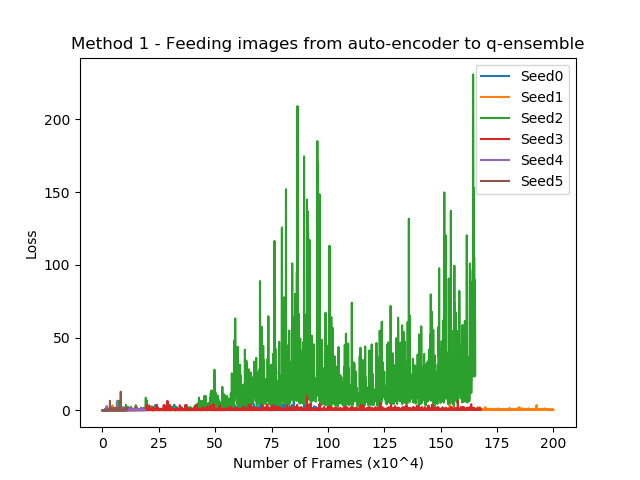}
  \captionof{figure}{Loss (Training)}
  \label{fig5b}
\end{subfigure}
\caption{Method 1 - Feeding auto-encoder images directly}
\label{fig5}
\end{figure}

Fig \ref{fig5} shows the reward and reward averaged per 100 instances for the combination of the encoder-decoder and q-ensemble models using different seeds. Fig \ref{fig5b} shows the loss for this method (Method 1). As can be seen, in comparison to just using the Q-ensemble, the combination hovers between the 400 and 800 mark for rewards without any improvement in training behavior even after 2.5 million timesteps. Also, training this model is really slow because of the multiple steps involved in action selection (generating next frames, feed to Q-ensemble, calculate path (if any) uncertainties). We restrict the path to be just the immediate action but repeat the action multiple times (4) to obtain a reasonable difference in the next states predicted by the auto-encoder.

In our analysis we found that the frames predicted by the auto-encoder are highly noisy themselves and this can lead to poor uncertainty estimates from the Q-ensemble (the Q-ensemble almost always has never seen such frames). This tends to bias the exploration in the wrong direction where actions with noisy predicted frames are always preferred instead of unexplored actions.
Our results for this method are quite similar to the results obtained by \cite{oh2015action}(Section 4.2) when they tried to replace the emulator with the frames prediction by the action-conditional model during testing.

\subsection{Method 2 - Trajectory Memory and Q-Ensembles}
\begin{figure}[ht]
\centering
\begin{subfigure}{0.6\textwidth}
  \centering
  \includegraphics[width=\textwidth]{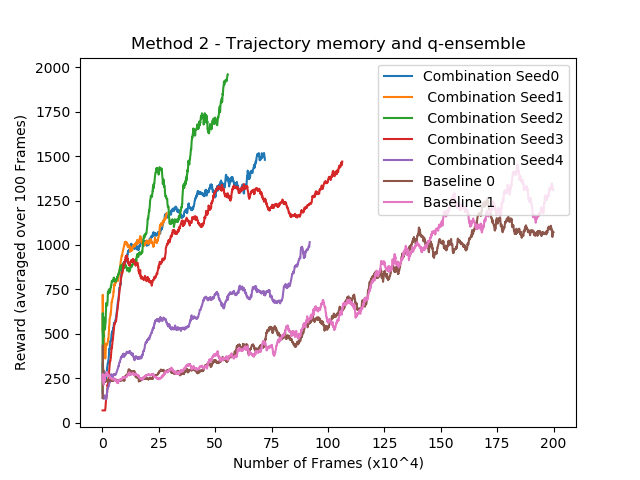}  
  \caption{Average Rewards (Every 100 episodes) (Training)}
  \label{fig:sub11}
\end{subfigure}%

\begin{subfigure}{0.6\textwidth}
  \centering
  \includegraphics[width=\textwidth]{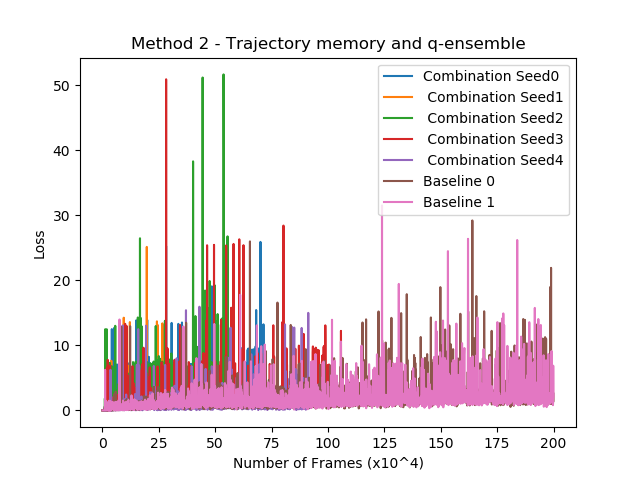}
  \captionof{figure}{Loss (Training)}
  \label{fig2b}
\end{subfigure}
\caption{Method 2 - Combining trajectory memory and q-ensemble estimates}
\label{fig2a}
\end{figure}

Figure \ref{fig:sub11} and \ref{fig2b} show the reward and loss plots obtained during training. As seen from the figures, the combination of trajectory memory and Q-ensemble variance (ucb-like exploration) often yields better rewards and reaches higher rewards very quickly ($<500K$ frames) compared to either of the baselines (Double DQN, Q-Ensembles). However, the behavior is highly dependent on the starting seeds and can lead to a lot of variance in performance (as seen in \ref{fig:sub11}). Training is again slow (not as slow as Method 1 though) and can take upto 12-15 hours to reach 1M timesteps. The results are encouraging and show that combining estimated visit frequency and variance in Q-estimates to drive exploration is much better than $\epsilon$-greedy or plain UCB-like exploration.

\section{Conclusion and Future Work}
Intelligent exploration strategies other than dithering ones like $\epsilon$-greedy are important for Q-learning in large state-spaces. We find that combining trajectory memory based visit estimates with variance estimates from a Q-ensemble improves exploration and helps the agent reach better rewards much faster than other methods. 

In the future, we hope to repeat these experiments on other Atari games like QBert or Seaquest where exploration is harder. As observed, the frames predicted by the auto-encoder are highly noisy themselves and this can lead to poor uncertainty estimates from the Q-ensemble. Using generator-discriminator methods that improve the quality of predicted future frames could be employed in the place of the encoder-decoder models. In order to address the stability issues generally faced by GANs, architectures like Wasserstein GANs will also be an important direction of research. With more realistic frames, it is easier to obtain unbiased uncertainty estimates from the Q-ensemble to drive exploration. Also, as exploration strategies become complex, training becomes really slow and it will become important to use computational tricks like separating exploration from Q-network training, parallel multiple environments etc. to reduce training time.

\FloatBarrier  
\bibliographystyle{unsrt}
\bibliography{ref}

\end{document}